\documentclass[10pt,twocolumn,letterpaper]{article}

\usepackage{3dv}
\usepackage{times}
\usepackage{epsfig}
\usepackage{graphicx}
\usepackage{amsmath}
\usepackage{amssymb}
\usepackage{multirow}

\usepackage[normalem]{ulem}
\DeclareMathOperator*{\argmin}{arg\,min}

\usepackage[pagebackref=true,breaklinks=true,letterpaper=true,colorlinks,bookmarks=false]{hyperref}

\threedvfinalcopy 

\def\threedvPaperID{146} 

\ifthreedvfinal\fi
\begin{document}

\title{Multi-Person 3D Human Pose Estimation from Monocular Images}

\author{Rishabh Dabral\\
IIT Bombay\\
{\tt\small rdabral@cse.iitb.ac.in}
\and
Nitesh B Gundavarapu\\
IIT Bombay\\
{\tt\small ntesh93@gmail.com}\\
\and
Rahul Mitra\\
IIT Bombay\\
{\tt\small rmitter@cse.iitb.ac.in}\\
\and
Abhishek Sharma\\
Axogyan AI\\
{\tt\small abhisharayiya@gmail.com}
\and
Ganesh Ramakrishnan\\
IIT Bombay\\
{\tt\small ganesh@cse.iitb.ac.in}
\and
Arjun Jain\\
Axogyan AI\\
{\tt\small arjunjain@gmail.com}
}

\maketitle
\begin{abstract}
Multi-person 3D human pose estimation from a single image is  a challenging problem, especially for in-the-wild settings due to the lack of 3D annotated data. We propose HG-RCNN, a Mask-RCNN based network that also leverages the benefits of the Hourglass architecture for multi-person 3D Human Pose Estimation. A two-staged approach is presented that first estimates the 2D keypoints in every Region of Interest (RoI) and then lifts the estimated keypoints to 3D. Finally, the estimated 3D poses are placed in camera-coordinates using weak-perspective projection assumption and joint optimization of focal length and root translations. The result is a simple and modular network for multi-person 3D human pose estimation that does not require any multi-person 3D pose dataset. Despite its simple formulation, HG-RCNN achieves the state-of-the-art results on MuPoTS-3D while also approximating the 3D pose in the camera-coordinate system.
\end{abstract}

\section{Introduction}

3D human pose-estimation consists of inferring the 3D joint-locations from an image or a sequence of images. It is the key to unlocking a large number of applications in AR/VR, Human-Computer-Interaction (HCI), Gaming, Activity Recognition, Surveillance, {\em etc.}. Although, there is a vast literature on single-person 3D pose estimation \cite{Sun_2017_ICCV,li20143d,zhou2016deep,zhou2016sparseness,Pavlakos_2017_CVPR,VNect_SIGGRAPH2017,Chen_2017_CVPR,yasin2016dual,bogo2016keep,DBLP:journals/tog/LoperM0PB15,akhter2015pose,Zhou_2017_ICCV,Dabral_2018_ECCV,humanMotionKanazawa19}, the space of multi-person 3D pose estimation is mostly unexplored with only a handful of prior work \cite{LCRNet,singleshotmultiperson2018,Zanfir_2018_CVPR,DBLP:LCRNet++,MubyNet}. Ironically, real-life human pose-estimation applications, most often, require multi-person pose estimation. For example, surveillance systems require real-time capturing of the poses for every person in the scene. Similarly, sports-analytics demands that all the players are simultaneously analyzed to capture inter-player interactions. Consequently, there exists a gap between existing research and real-world requirements. 

A simple extension of the single-person pose estimation systems to the multi-person setting involves separate detection of every person followed by single-person pose estimation on person crop. 

Unfortunately, the run-time of this approach is likely to increase linearly with the number of people in the scene, making it inefficient for analysis in crowded scenes. Additionally, most existing multi-person pose estimation methods~\cite{LCRNet,singleshotmultiperson2018,DBLP:LCRNet++}, with the exception of~\cite{Zanfir_2018_CVPR} estimate 3D pose configuration only relative to the root joint. However, relative spatial ordering of different people in the scene is also needed to facilitate reasoning about human interactions and provide a better understanding of the scene. Relative spatial estimation has the potential to unlock accurate tracking of multiple persons in a scene video. 

Moreover, most prior work on multi-person pose estimation~\cite{LCRNet,singleshotmultiperson2018,MubyNet} relies on creating or simulating a multi-person 3D human pose dataset as a necessity for training. The pre-requisite is due to the end-to-end integrated person detection and pose estimation pipeline. This limits the variability presented to the system while training because obtaining real-world in-the-wild 3D annotations in multi-person setting is challenging, expensive and a research problem in itself. 

\begin{figure*}[t] 
	\centering
	\includegraphics[width=1\linewidth]{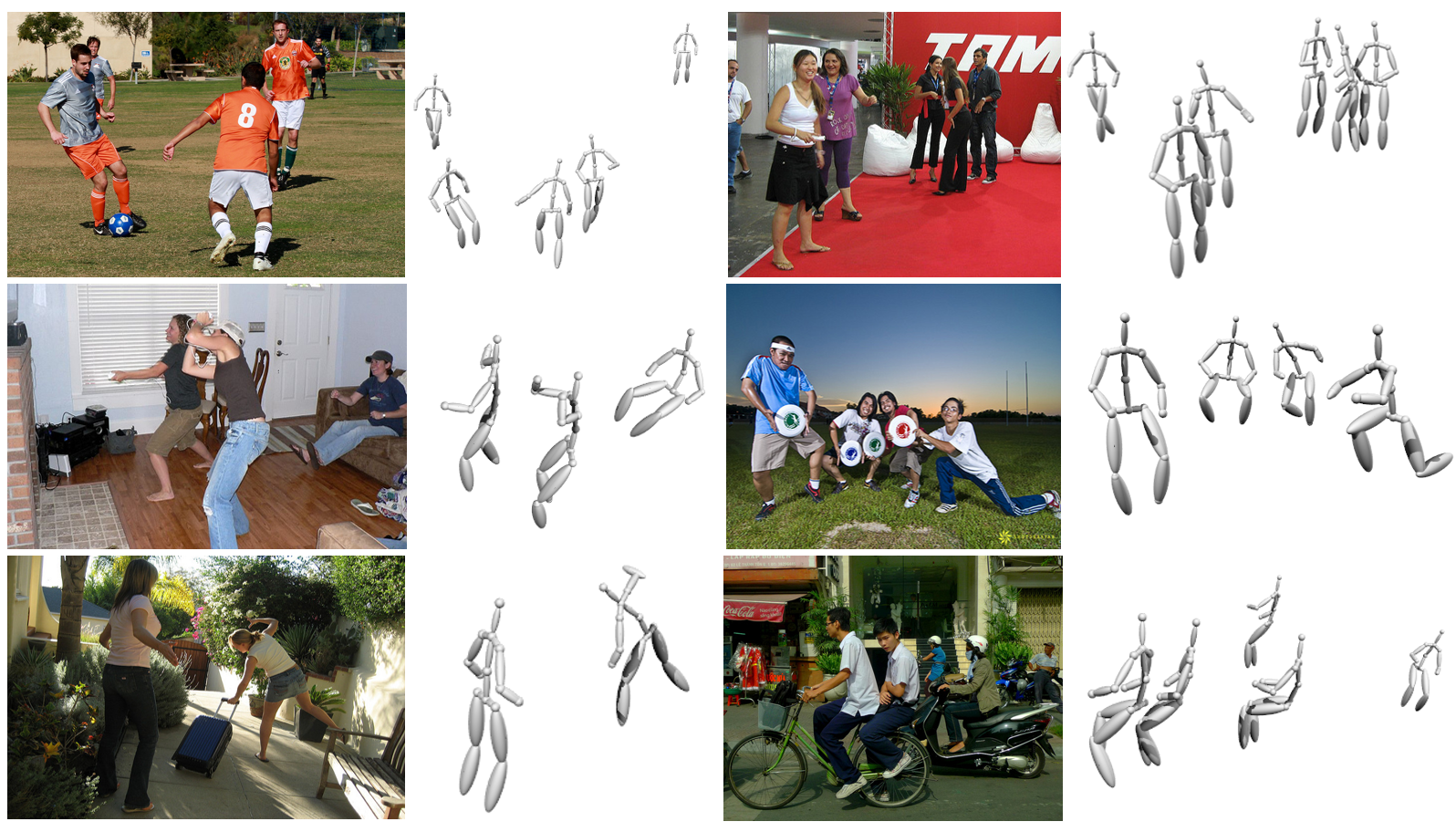}
    \caption{Some results of our proposed 3D pose estimation pipeline on some challenging samples from MS COCO. Our approach is resilient against occlusions and clutter. We also approximate the spatial ordering of people in the scene with respect to the camera. Further in-the-wild results and a 3D rendered view of the above images can be found in the supplementary material.} 
	\vspace{-1em}
	\label{fig:MuPots}
\end{figure*}

In light of the aforementioned discussion of multi-person 3D pose estimation, we propose a quasi top-down architecture that decouples the 2D key-point detection and 2D-to-3D lifting tasks. The proposed architecture, HG-RCNN, brings together the goodness of Mask-RCNN~\cite{maskrcnn} and the Hourglass~\cite{NewellYD16} network for heatmap regression. The regressed heatmaps are then fed to an independently trained lifting module to regress the root-relative 3D poses. Consequently, we completely avoid using any multi-person 3D pose dataset in the pipeline since it leverages the existing multi-person 2D pose datasets and single-person 3D pose datasets. Owing to its modular architecture, the first step of obtaining 2D poses can be trained with publicly available large-scale in-the-wild multi-person datasets, such as COCO~\cite{MSCOCO:2014}, LIP~\cite{LIP} and MPII 2D dataset~\cite{andriluka14cvpr}. This allows HG-RCNN to cope with challenging variations in view-point, lighting, apparel, occlusion and extreme poses without the need of costly 3D annotations in-the-wild setting. The keypoint heatmaps from the HG-RCNN are passed through a \textit{soft-argmax} module and fed to a 2D-3D lifting module. Finally, our pipeline approximates pose-configurations in camera coordinates without the need of costly geometric optimization. The resulting system outperforms all previous approaches on the challenging MuPoTS-3D~\cite{singleshotmultiperson2018} test-set that contains a majority of in-the-wild test scenarios. The method generalizes well to in-the-wild images, even without exploiting any structural priors, while running at 12-15fps on images of size $400\times600$ on a single Nvidia 1080Ti graphics card. 

In summary, we contribute a state-of-the-art model for performing in-the-wild multi-person 3D pose estimation. The model can be trained without using any multi-person 3D dataset and the system also estimates the relative ordering of the persons in the 3D space.

\section{Related Works}

Human Pose Estimation has been a widely studied problem. Here, we describe prior art relevant to this work from three broad viewpoints: (a) 2D Pose estimation, (b) Single-person 3D Pose estimation and (c) Multi-Person 3D Pose estimation. A detailed survey of the area can be found in ~\cite{SARAFIANOS20161}.

\indent \textbf{2D Human Pose Estimation:} Most 2D human pose estimation methods represent their joint outputs as heatmaps, wherein a heatmap's value at a point represents the possibility of the corresponding joint's existence in that position. \cite{Wei2016ConvolutionalPM} proposed Convolutional Pose Machines that iteratively refined the heatmap predictions at every stage. The Stacked Hourglass network~\cite{NewellYD16} was an encoder-decoder architecture with skip connections to facilitate joint reasoning of high level structural and low level textural features of human pose. Mask-RCNN~\cite{maskrcnn} proposed an extension of Faster-RCNN~\cite{ren2015faster} for simultaneously predicting the pose and 2D keypoints and/or instance segmentation masks. In a similar line of work, ~\cite{DensePose} predicted the \textit{u-v} maps of the persons which can then be used for dense reconstruction. \cite{Sekii_2018_ECCV} proposed a variant to Mask-RCNN by defining joints as regions instead of persons. In similar spirits, our proposed pipeline attempts to synergise Mask-RCNN and Hourglass networks for multi-person 3D pose estimation task.

\textbf{Single-person 3D Pose Estimation:} Single person 3D pose estimation works can be broadly divided based on whether they directly regress 3D joints~\cite{Sun_2017_ICCV,li20143d,zhou2016deep,zhou2016sparseness} or use a pipelined approach of inferring 3D pose from 2D pose~\cite{Tome_2017_CVPR,Zhou_2017_ICCV,VNect_SIGGRAPH2017,Moreno-Noguer_2017_CVPR,Lin_2017_CVPR}. VNect~\cite{VNect_SIGGRAPH2017} proposed the first real-time approach and parameterized a 3D joint by a heatmap and 3 location maps. Using a 2D-to-3D pipeline enables the use of rich 2D pose datasets which, in turn, improves in-the-wild generalizability. Many approaches perform a direct 2D-to-3D lifting of poses ~\cite{zhou2016sparseness,martinez2017,Moreno-Noguer_2017_CVPR,Chen_2017_CVPR,yasin2016dual} by either learning the transformation or by a nearest-neighbour lookup in a pose library. Furthermore, many pipelined approaches~\cite{VNect_SIGGRAPH2017,LCRNet,Zhou_2017_ICCV,Sun_2017_ICCV,zhou2016sparseness,Pavlakos_2017_CVPR} have reported significant improvements in in-the-wild performances by using the more diverse 2D pose datasets to pre-train or jointly train their 2D prediction modules.

Several methods in the past have also reported significant improvements by using temporal cues~\cite{Hossain_2018_ECCV,VNect_SIGGRAPH2017,zhou2016sparseness,Dabral_2018_ECCV,Nie_2017_ICCV,Zanfir_2018_CVPR} by either learning a motion/refinement model or by using temporal constraints in a constrained optimization framework. 

\begin{figure*}[!h] 
	\centering
	\includegraphics[width=0.95\textwidth, height=7cm]{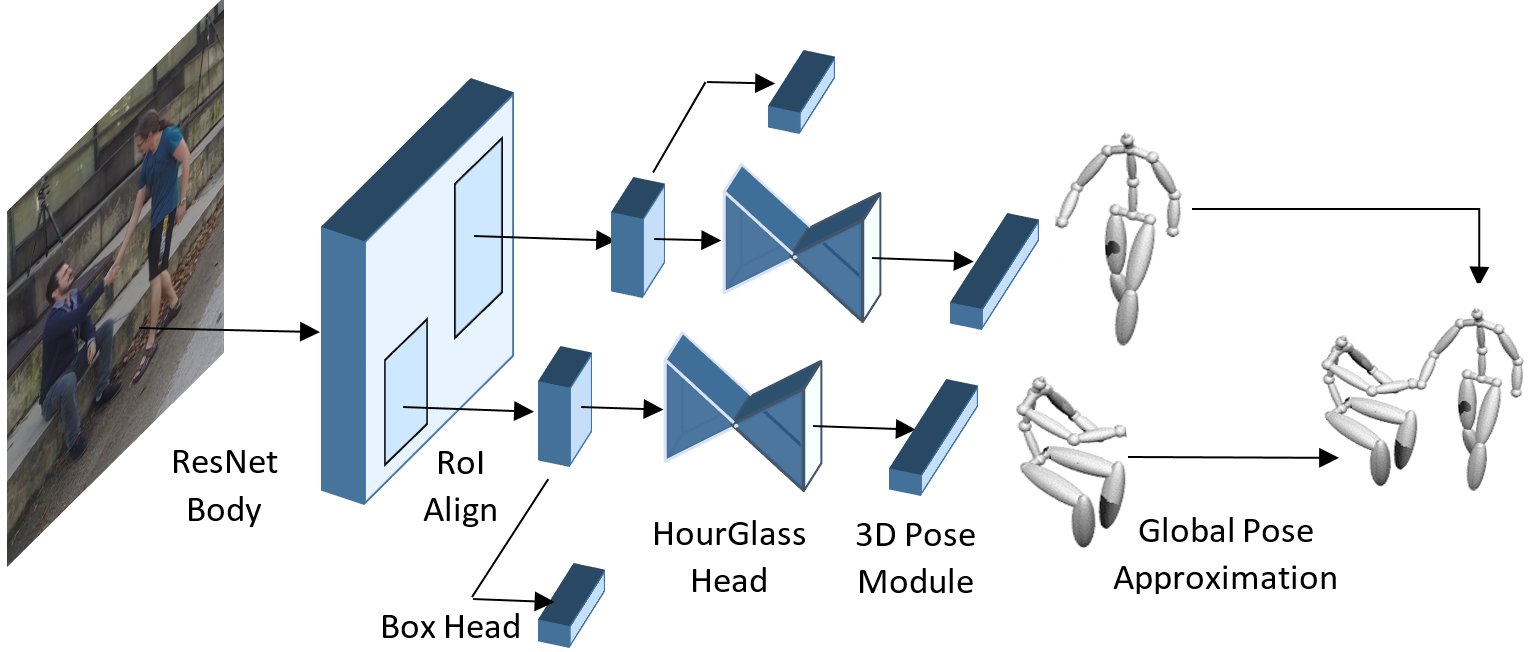}
    \caption{Schematic of our Multi-Person 3D Pose Estimation approach. We augment the Faster-RCNN \cite{ren2015faster} architecture with a shallow HourGlass Network \cite{NewellYD16}. The heatmaps generated by the hourglass are then input to a 3D Pose Module which regresses the root-relative 3D joint coordinates. The estimated 3D poses of all the Regions of Interests (RoI's) are then collected and their global root positions are approximated to ensure that relative spatial ordering is preserved. } 
	\vspace{-1em}
	\label{fig:HGRCNN}
\end{figure*}

\indent \textbf{Multi-Person 3D Pose Estimation:}
Broadly, multi-person pose estimation approaches, 2D and 3D alike, can be classified into top-down and bottom-up approaches. Bottom-up approaches simultaneously predict all the key-points followed by assembling them into full poses for all persons. On the other hand, top-down approaches first detect the human candidates and subsequently perform pose estimation for each of them. While bottom-up methods are lucrative in terms of efficiency, they tend to be less accurate. For example, the top 5 entries in MS COCO key-points challenge employ top-down approaches~\cite{MSCOCO:2014}. Intuitively, it makes sense to solve for pose estimation on a person's crop, instead of solving a much more challenging problem of grouping detected key-points into a full person. In recent years, however, a middle ground has been found in the form of quasi top-down architectures based on Mask-RCNN~\cite{maskrcnn,DensePose,hu2017everything,Sekii_2018_ECCV} that have been successful in simultaneously detecting the object RoIs and performing downstream tasks on the corresponding RoI feature-maps, without having to crop the image back.

LCRNet~\cite{LCRNet} was the first method to perform Multi-Person 3D Pose Estimation. They propose an integrated network based on Faster-RCNN~\cite{ren2015faster} which first proposes Regions of Interest (RoIs) that are fed to a classifier and a regressor. The classifier estimates the most probable anchor pose out of the $K$ pre-defined anchor poses obtained from a MoCap dataset. The regressor then refines the anchor poses towards an accurate pose prediction. Alternately, ~\cite{singleshotmultiperson2018} propose a bottom-up approach wherein they regress the heatmaps along with X, Y, and Z location maps for every image. The location maps provide the corresponding 3D positions of joints in metric space. The estimated 3D joints are then associated using Part Affinity fields~\cite{cao2017realtime} based on the heatmaps. Both the approaches depend on the explicit creation or simulation of  multi-person 3D pose datasets for training. Our method, on the other hand, avoids the use of such datasets and relies on 3D data only for the single person case. Further, Zanfir \etal ~\cite{Zanfir_2018_CVPR} proposed a large-scale human sensing system for multiple people that estimates pose and shape using the top-down approach of person detection followed by pose estimation for each person. Recently, Zanfir \etal ~\cite{MubyNet} proposed MubyNet, a bottom-up approach that performs joint association by formulating it as a binary integer programming problem.  In contrast, Mask-RCNN \cite{maskrcnn} based quasi top-down methods~\cite{DensePose,hu2017everything,LCRNet} have proven to be effective for simultaneously locating objects at a coarse level and detecting finer spatial layouts like segmentation masks, key-point heatmaps, u-v maps, etc.. Our proposed HG-RCNN exploits this setting and also regresses for 3D key-points. However, unlike LCRNet~\cite{LCRNet} and LCRNet++~\cite{DBLP:LCRNet++}, our method does not require anchor-poses and is relatively simpler.

\section{Problem Formulation}
Given an image $I$ containing $N$ people, we estimate the poses $P = (P_1, P_2, ..., P_N)$, wherein $P_i \in \mathcal{R}^{n\times3}$ and $n$ is the number of joints. Every pose $P_i$ is a set of $n$ joints in 3D Euclidean space with the origin set to a root joint, pelvis in this case. As an intermediate step, our method first estimates the 2D key-points $K = (K_1, K_2, ..., K_N)$ with $K_i \in \mathcal{R}^{n\times2}$ in the image coordinate space. Finally, we approximate the global poses $P^G = (P^G_1, P^G_2, ..., P^G_N)$ in camera coordinate space. 

\subsection{Multi-Person 3D Pose Estimation}
We follow a generic, two-step pipeline for root-relative 3D pose estimation. First, we estimate per-frame 2D key-points of all the people in an image and lift them to 3D pose using a simple residual network. We use a Mask-RCNN based architecture to estimate 2D key-points. However, vanilla keypoint head of Mask-RCNN is not the most conducive architecture for reasoning with structured/articulated objects like human pose. Fortunately, the Hourglass~\cite{NewellYD16} family of networks have been found to be extremely effective in reasoning about a human pose in a structure-aware way. Therefore, we propose to employ a tiny Hourglass head as a surrogate to the key-point head. This simple patch alone leads to noticeable improvements in the results and will be discussed further in Section 5.2. 

In the second step, the obtained keypoint heatmaps are lifted to 3D joints using a network with two residual modules of size 2048. When deployed in wild settings, it is trained with the heatmaps regressed on the MPI-INF-3DHP training dataset~\cite{mono-3dhp2017} which provides a wide variety of viewpoints and poses activities, thereby adding to the generalization capability of the network. It is worth noting that it is this modular structure of the pipeline that allows us to train the network without any multi-person 3D dataset. The in-the-wild performance is guaranteed by two aspects: a) The heatmaps are learnt on completely wild multi-person 2D keypoint datasets, and b) the lifting module is agnostic to the image features and trained on a dataset consisting of a wide variety of 2D-3D paired annotations. 

Further details on the architecture are discussed in Section 4. At this stage, all the outputs (3D keypoints) are in their individual root relative space. For placing the detected poses in camera-relative space, we estimate the common focal length of the camera and the translation vectors from the individuals' roots to the camera center.

\subsection{Global Pose Approximation}
Our approach for camera-relative pose approximation is based on jointly optimizing the root joints' global positions and the camera's focal length for the projection error. We initialize the root joint positions using a\textit{ weak-perspective} projection assumption, thus, requiring us to estimate the shrinking parameter $\mathbf{\alpha_i}$ for every pose $Pi$ in the scene. To this end, we compute the sum of bone lengths of the 2D keypoints, $S_{2D}$, followed by computing the sum of bone lengths, $S_{3D}$, of the 3D pose's orthographic projection. 

The ratio $S_{2D}/S_{3D}$ acts as a surrogate to the shrinking factor $\mathbf{\alpha_i}$. This finally leads to the following formulation for estimating the global $X$ (horizontal) and $Z$ (depth) coordinates of a joint:
        \begin{equation}
            Z = f * \frac{S_{3D}}{S_{2D}},
        \end{equation}
        \begin{equation}
            X = (x-o_x) * \frac{S_{3D}}{S_{2D}}
        \end{equation}
where, $x$ corresponds to the 2D keypoint and $o_x$ is the $x$ co-ordinate of the image center. The focal length, $f$, is initialized by assuming a field-of-view of $60^\circ$. The same formulation holds for the $Y$ (vertical) coordinate as well.

Once the root translations are initialized and the full 3D poses are placed in the respective root positions, we iteratively optimize the translation and focal length. The global rotations are assumed to be identity. Thus, the objective function can be written as:
\begin{equation}
   f^{*}, t^{*} =  \argmin_{f,t} \sum_{i=1}^N \vert \vert K_i - \Pi_{f,t_i} P_i \vert \vert_2
\end{equation}
where $t = \{t_1, t_2, \dots t_N\}$ with $t_i$ being the translation vector of $i^{th}$ subject's root joint and $\Pi$ being the projection operator. This, finally, leads to the global pose, $P_i^G = P_i + t_i^{*}$.

It is worth noting that the proposed global pose approximation method is just an approximation that can be quickly implemented and run in real-time. The approximation is not expected to work when the person is aligned with the optical axis. We discuss further limitations in section 6. It is not intended to be highly accurate, but only expected to make spatial ordering apparent to systems that need it, eg. action recognition. 

\begin{table*}[t]
\setlength\tabcolsep{1.2pt}
\caption{Comparison of our method with prior work on MuPoTS-3D on \textit{Setting 1}. The \textbf{top half} shows results on \textit{all annotated poses} in the test set. The \textbf{bottom half} shows results when only the detected poses are considered. The evaluation metric is 3D PCK and higher is better. *Note, that the average PCK provided in LCRNet++~\cite{DBLP:LCRNet++} is not weighed by the number of persons in each test sequence unlike~\cite{LCRNet,singleshotmultiperson2018} and ours.} \label{tab:mupots_full_p1}
\vskip 2mm
\begin{tabular}{ l | c  c  c  c  c  c  c  c  c  c  c  c  c  c  c  c  c  c  c  c  c}
\hline
Method & TS1 & TS2 & TS3 & TS4 & TS5 & TS6 & TS7 & TS8 & TS9 & TS10 & TS11 & TS12 & TS13 & TS14 & TS15 & TS16 & TS17 & TS18 & TS19 & TS20 & Avg\\
\hline
\cite{LCRNet} & 67.7 & 49.8 & 53.4 & 59.1 & 67.5 & 22.8 & 43.7 & 49.9 & 31.1 & 78.1 & 50.2 & 51.0 & 51.6 & 49.3 & 56.2 & 66.5 & 65.2 & 62.9 & 66.1 & 59.1 & 53.8\\
\cite{singleshotmultiperson2018} & 81.0 & 59.9 & 64.4 & 62.8 & 68.0 & 30.3 & 65.0 & 59.2 & {64.1} & {83.9} & 67.2 & 68.3 & 60.6 & {56.5} & {69.9} & 79.4 & 79.6 & 66.1 & 66.3 & 63.5 & 65.0 \\
\cite{DBLP:LCRNet++}* & {87.3} & 61.9 & 67.9 & 74.6 & \textbf{78.8} & 48.9 & 58.3 & 59.7 & \textbf{78.1} & \textbf{89.5} & 69.2 & \textbf{73.8} & \textbf{66.2} & 56.0 & \textbf{74.1} & 82.1 & 78.1 & 72.6 & 73.1 & 61.0 & 70.6 \\
\cite{xnect} & \textbf{88.4} & 65.1 & 68.2 & 72.5 & 76.2 & 46.2 & \textbf{65.8} & \textbf{64.1} & 75.1 & 82.4 & 74.1 & 72.4 & 64.4 & \textbf{58.8} & 73.7 & {80.4} & \textbf{84.3} & 67.2 & 74.3 & 67.8 & 70.4\\
Ours & {85.1} & \textbf{67.9} & \textbf{73.5} & \textbf{76.2} & 74.9 & \textbf{52.5} & {65.7} & {63.6} & 56.3 & 77.8 & \textbf{76.4} & {70.1} & {65.3} & 51.7 & 69.5 & \textbf{87.0} & {82.1} & \textbf{80.3} & \textbf{78.5} & \textbf{70.7} & \textbf{71.3}\\
\hline
\cite{LCRNet} & 69.1 & 67.3 & 54.6 & 61.7 & 74.5 & 25.2 & 48.4 & 63.3 &  {69.0} & 78.1 & 53.8 & 52.2 & 60.5 & 60.9 & 59.1 & 70.5 & 76.0 & 70.0 & 77.1 & 81.4 & 62.4\\
\cite{singleshotmultiperson2018} & 81.0 & 64.3 &  {64.6} &  {63.7} & 73.8 & 30.3 & 65.1 & 60.7 & 64.1 &  {83.9} & 71.5 & 69.6 & 69.0 & {69.6} &  {71.1} & 82.9 & 79.6 & 72.2 & 76.2 & 85.9 & 69.8\\
\cite{DBLP:LCRNet++}* & {88.0} & 73.3 & {67.9} & \textbf{74.6} & {81.8} & 50.1 & 60.6 & 60.8 & \textbf{78.2} & \textbf{89.5} & 70.8 & \textbf{74.4} & \textbf{72.8} & 64.5 & \textbf{74.2} & 84.9 & 85.2 & 78.4 & 75.8 & 74.4 & 74.0 \\
\cite{xnect} & \textbf{88.4} & 70.4 & \textbf{68.3} & 73.6 & \textbf{82.4} & 46.4 & 66.1 & \textbf{83.4} & 75.1 & 82.4 & 76.5 & 73.0 & 72.4 & \textbf{73.8} & 74.0 & 83.6 & 84.3 & 73.9 & \textbf{85.7} & 90.6 & \textbf{75.8} \\
Ours & {85.8} & \textbf{73.6} & 61.1 & 55.7 &  {77.9} & \textbf{53.3} & \textbf{75.1} & {65.5} & 54.2 & 81.3 & \textbf{82.2} & {71.0} &  {70.1} & 67.7 & 69.9 & \textbf{90.5} & \textbf{85.7} & \textbf{86.3} & {85.0} & \textbf{91.4} & {74.2}\\
\hline
\end{tabular}
\vskip 2mm
\vspace{-1em}
\end{table*}
\begin{table*}[t]
\setlength\tabcolsep{1pt}
\caption{Performance of our method on MuPoTS using the \textit{Setting 2}. The \textbf{top half} shows results on all annotated poses in the test set. The \textbf{bottom half} shows results when only the detected poses are considered. 'all' corresponds to evaluation on all eligible persons and 'occ' corresponds to the results on occluded persons. The evaluation metric is 3D PCK and higher is better. Notice that compared to \ref{tab:mupots_full_p1}, the improvement is mostly observed on sequences with significant occlusion, eg. TS18 and TS19. }\label{tab:mupots_full_p2}
\vskip 2mm
\begin{tabular}{ l | c  c  c  c  c  c  c  c  c  c  c  c  c  c  c  c  c  c  c  c  c}
\hline
Method & TS1 & TS2 & TS3 & TS4 & TS5 & TS6 & TS7 & TS8 & TS9 & TS10 & TS11 & TS12 & TS13 & TS14 & TS15 & TS16 & TS17 & TS18 & TS19 & TS20 & Avg\\
\hline
Ours (occ) & 82.0 & 61.1 & 62.3 & 70.2 & 62.7 & 53.3 & 67.8 & 63.1 & 59.8 & 39.1 & 73.1 & 69.3 & 67.4 & 33.7 & 59.0 & 79.1 & 79.0 & 82.5 & 79.7 & 36.2 & 64.0\\
Ours (all) & 85.2 & 69.2 & 73.1 & 75.6 & 77.7 & 52.1 & 65.1 & 66.0 & 57.7 & 77.6 & 76.6 & 69.0 & 71.6 & 53.4 & 70.3 & 86.0 & 84.3 & 84.7 & 83.7 & 72.8 & 72.6\\
\hline
Ours (occ) & 82.0 & 61.1 & 62.3 & 71.0 & 62.7 & 53.8 & 69.0 & 63.2 & 59.8 & 39.1 & 78.0 & 69.8 & 67.4 & 47.4 & 59.0 & 79.2 & 79.5 & 82.5 & 79.7 & 76.6 & 67.1\\ 
Ours (all)& 85.2 & 69.2 & 73.1 & 76.1 & 77.7 & 52.7 & 65.9 & 66.0 & 57.7 & 77.6 & 80.5 & 69.1 & 71.6 & 60.6 & 70.3 & 87.1 & 85.1 & 84.7 & 83.7 & 92.1 & 74.3 \\
\hline
\end{tabular}
\vspace{-1em}
\end{table*}

\section{Network and Training Details}
\textbf{HG-RCNN:} The HG-RCNN is constructed by appending an hourglass on the keypoint head of Mask RCNN as shown in Figure (\ref{fig:HGRCNN}). Instead of upsampling once while deconvolving and once at the final layer, we upsample (with $4 \times$) the feature maps all at once before passing the feature values on to the hourglass. The number of feature-maps is brought down from $512$ to $128$ using a $1 \times 1$ convolution layer. The original hourglass is modified to have three nested residuals (instead of $4$) and has a feature-map of size $7 \times 7$ at the bottle-neck layer. The hourglass output is then fed to a final classification layer which predicts the heatmaps for every joint. 

We train the network described above with the Cross-Entropy Loss. While finetuning, we train on top 500 RoIs and use a batch size of 16. The network is trained with a base learning rate of $0.02$ on a single Nvidia P6000 Quadro graphics card.  

\textbf{3D Pose Module:} Our 2D-to-3D pose module converts the heatmap activations to 3D pose using a residual architecture and is in line with the 2D-3D lifting pipelines proposed in ~\cite{martinez2017, Sun_2018_ECCV, Moreno-Noguer_2017_CVPR}. 
We input the 2D poses in heatmap space after passing the heatmaps through a \textit{softargmax} layer. This has two benefits: a) it makes learning possible from images of any given size and scale, and b) it facilitates end-to-end training of the network architecture. The network is trained using RMSProp optimizer and a learning rate of $2.5\exp{-4}$ which is reduced by 10 times after 40 epochs. 

While testing on MuPoTS (multi-person) dataset, we use the 3D pose module trained only on MPI-INF-3DHP dataset because both the training and the test sets had the same motion capture system. Human3.6 was captured by a different mocap system which leads to the same joint name pointing to different physical locations on the body. 

\section{Experiments}
This section describes our experiments on MuPoTS-3D ~\cite{singleshotmultiperson2018}, MS COCO ~\cite{MSCOCO:2014} and Human3.6M ~\cite{h36m_pami} datasets.
\subsection{Evaluation Datasets}

\begin{figure*}[!h] 
	\centering
	\includegraphics[width=1\linewidth]{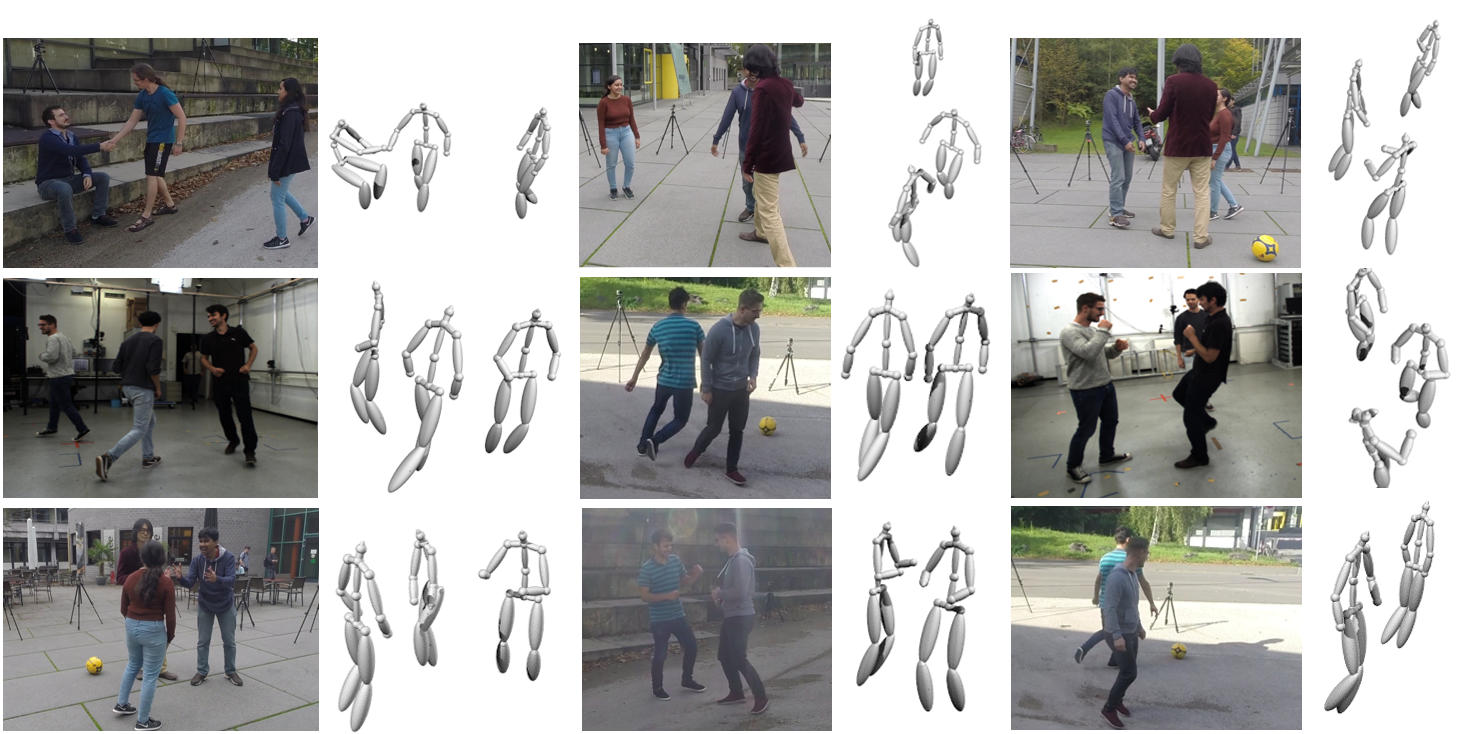}
    \caption{Visualization of our results on MuPoTS-3D Test Set from different viewpoints. Notice that the model is fairly robust to occlusions. The spatial alignment is not derived from ground truth.} 
	\vspace{-1em}
\end{figure*}
\textbf{MuPoTS-3D Test Set:} Multi-Person Test Set 3D ~\cite{singleshotmultiperson2018} is a recently released \textit{multi-person} 3D human pose test dataset. It consists of 20 test sequences shot with a marker-less mocap system - 5 indoor and 15 outdoor. Every sequence contains 2-3 persons in a variety of activities. The evaluation metric used is 3D PCK - percentage of correct keypoints within a radius of 15cm - on all the annotated persons. In case of a missed detection, all the joints of the missed person are considered erroneous. An alternative evaluation mode is the one in which the evaluations are performed only on the detected joints. 

The official evaluation code performs a greedy matching of detections and ground truth based on the number of 2D keypoints within a proximity of $40px$. We call this method \textit{Setting 1} for MuPoTS. 

We also evaluate our model in the setting wherein the greedy matching is done based on 3D distances instead of 2D distances. We call this \textit{Setting 2}. This joint matching strategy is, arguably, less sensitive to cases of heavy occlusion which would, otherwise, confuse a keypoint based matching detector. This, as discussed in Section 5.2, leads to missed detections even when the model actually detects the appropriate person. Note, that the two settings differ only in the way the predicted poses are matched with the ground truth poses. All the other details of evaluation, like 3D PCK threshold, joints used for matching, etc remains the same. 

\begin{figure*}[t] 
	\centering
	\includegraphics[width=1\linewidth]{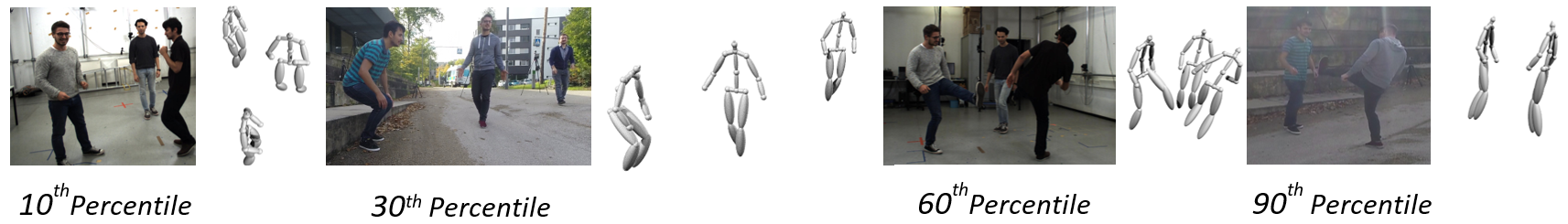}
    \caption{MPJPE based Percentile Analysis on MuPoTS-3D test set. The lower percentile is better. An important inference from the analysis is that the method is sensitive to lighting and low contrast setting.} 
	\vspace{-1em}
	\label{fig:Percentile}
\end{figure*}

\begin{table*}[t]
\centering
\caption{Comparative evaluation of our model on Human 3.6 using Absolute MPJPE. The evaluations were performed on subjects 9 and 11. The papers above the horizontal line are single-person pose estimation papers and the ones below the line are multi-person pose estimation papers.} \label{tab: h36mp1}
\begin{tabular}{l  c  c  c  c  c  c  c  c }
\hline
Method & Direction & Discuss & Eat & Greet & Phone & Pose & Purchase & Sit \\
\hline
\hline
Martinez~\cite{martinez2017} &  51.8   & 56.2 & 58.1 & 59.0 & 69.5 & 55.2 & 58.1 & 74.0  \\ 
Zhou~\cite{Zhou_2017_ICCV} & 54.8 & 60.7 & 58.2 & 71.4 & 62.0 & 53.8 & 55.6 & 75.2 \\
Sun~\cite{Sun_2017_ICCV} & 52.8 & 54.8 & 54.2 & 54.3 & 61.8 & 53.1 & 53.6 & 71.7 \\
Dabral~\cite{Dabral_2018_ECCV} & 44.8 & 50.4 & 44.7 & 49.0 & 52.9 & 43.5 & 45.5 & 63.1 \\
Hossain~\cite{Hossain_2018_ECCV} & 44.2 & 46.7 & 52.3 & 49.3 & 59.9 & 47.5 & 46.2 & 59.9 \\
Sun~\cite{Sun_2018_ECCV} & 47.5 & 47.7 & 49.5 & 50.2 & 51.4 & 43.8 & 46.4 & 58.9 \\
\hline
Rogez~\cite{LCRNet} & 76.2 & 80.2 & 75.8 & 83.3 & 92.2 & 79.0 & 71.7 & 105.9 \\ 
Mehta~\cite{singleshotmultiperson2018} & 58.2 & 67.3 & 61.2 & 65.7 & 75.8 & 62.2 & 64.6 & 82.0 \\
Rogez~\cite{DBLP:LCRNet++} & 50.9 & 55.9 & 63.3 & 56.0 & 65.1  & 52.1 & 51.9 & 81.1 \\
Ours (Baseline) & 60.2 & 64.5 & 66.2 & 70.1 & 75.6 & 65.4 & 69.4 & 83.7 \\
Ours (Fine-Tuned) & 52.6 & 61.0 & 58.8 & 61.0 & 69.5 & 58.8 & 57.2 & 76.0 \\
\hline
\hline
Method & SitDown & Smoke & Photo & Wait & Walk & WalkDog & WalkPair & Avg \\
\hline
\hline
Martinez~\cite{martinez2017}& 94.6 & 62.3 & 78.4 & 59.1&65.1&49.5&52.4&62.9 \\
Zhou~\cite{Zhou_2017_ICCV}& 111.6 & 64.1 & 65.5 & 66.0 & 51.4 & 63.2 & 55.3 & 64.9 \\
Sun~\cite{Sun_2017_ICCV} & 86.7 & 61.5 & 67.2 & 53.4 & 47.1 & 61.6 & 53.4 & 59.1 \\
Dabral~\cite{Dabral_2018_ECCV} & 87.3 & 51.7 & 61.4 & 48.5 & 37.6 & 52.2 & 41.9 & 52.1 \\
Hossain~\cite{Hossain_2018_ECCV} & 65.6 & 55.8 & 59.4 & 50.4 & 52.3 & 43.5 & 45.1 & 51.9 \\
Sun~\cite{Sun_2018_ECCV} & 65.7 & 49.4 & 55.8 & 47.8 & 38.9 & 49.0 & 43.8 & 49.6 \\
\hline
Rogez~\cite{LCRNet} & 127.1 & 88.0 & 105.7 & 83.7 & 64.9 & 86.6 & 84.0 & 87.7 \\
Mehta~\cite{singleshotmultiperson2018} & 93.0 & 68.8 & 84.5 & 65.1 & 57.6 & 72.0 & 63.6 & 69.9\\
Rogez~\cite{DBLP:LCRNet++}&  91.7 & 64.7 & 70.7 & 54.6 & 44.7 & 61.1 & 53.7 & \textbf{61.2} \\
Ours (Baseline) & 105.7 & 70.2 & 89.6 & 69.1 & 61.7 & 80.6 & 66.9 & 73.0 \\
Ours (Fine-Tuned) & 93.6 & 63.1  & 79.3 & 63.9 & 51.5 & 71.4 & 53.5 & 65.2 \\

\hline
\end{tabular}
\vspace{-1em}
\end{table*}

\textbf{Human3.6:} Human 3.6M~\cite{h36m_pami} is a single-person 3D human pose dataset captured with marker-based motion capture system. It consists of $11$ subjects performing $15$ actions. We evaluate our model on the commonly followed protocol~\cite{VNect_SIGGRAPH2017,Sun_2017_ICCV,Zhou_2017_ICCV,LCRNet,mono-3dhp2017,Dabral_2018_ECCV,Moreno-Noguer_2017_CVPR} that uses subjects $1,5,6,7$ and $8$ for training, The evaluations are done on subjects $9$ and $11$. All the videos are downsampled from $50 fps$ to $10 fps$. The evaluation metric used is Mean Per Joint Position Error (MPJPE) which is calculated after aligning only the roots of the predicted and ground truth 3D poses. 

\textbf{MSCOCO Keypoints:} MSCOCO Keypoints is a large scale dataset for 2D multi-person keypoint detection task with roughly 110k training images. It also provides the person bounding boxes and segmentation masks. The 2D keypoint detection task is evaluated on the commonly used Average Precision (AP) metric at different threshold levels. Similarly, the quality of bounding box detections are evaluated using AP.

\subsection{Quantitative Evaluation}
We now discuss the numerical results achieved on the datasets mentioned above.

\textbf{MuPoTS-3D Test Set:} Table~\ref{tab:mupots_full_p1} compares the performance of our simple yet effective method with the existing multi-person 3D pose results. On \textit{Setting 1}, we improve the state-of-the art significantly with a 3DPCK of 71.25\% as against 65\% in \cite{singleshotmultiperson2018} and $53.8\%$ in \cite{LCRNet}. For LCRNet \cite{LCRNet}, the reported results are evaluated by \cite{singleshotmultiperson2018}. We report an improved performance on several test sequences. We also significantly improve the performance of occluded joints ($61\%$ vs $48.7\%$) as well as the non-occluded joints ($75.6\%$ vs $70\%$) when compared with~\cite{singleshotmultiperson2018}. Our method also performs significantly well when only detected persons are compared. In this setting, we observe $75\%$ 3DPCK while the state-of-the-art being $69.8\%$. We also compare our performance with the recently released XNect~\cite{xnect} and demonstrate competitive results on all annotated poses (71.3\% vs 70.4\%) as well as the detected poses (74.2\% vs 75.8\%).

We also evaluate our method on the proposed \textit{Setting 2}. We observed an improved 3DPCK of $72.6\%$ when compared with \textit{Setting 1}. This improvement is facilitated by a simple tweak in the greedy matching algorithm of ground-truth and predicted persons. On deeper inspection, we see sharp improvements in sequences with heavy occlusions, like TS18 and TS19. Further, the overall improvement is significant when comparing the performance of occluded joints ($64\%$ vs $61\%$ of~\cite{singleshotmultiperson2018}). This observation can be attributed to the fact that matching predictions with ground-truths based on 2D keypoints leads to matching errors and missed detections when two or more persons occlude each other.  Indeed, we observe that the algorithm's detection percentage rose from 93\% to 96\%, thus improving the overall 3DPCK.  Interestingly, we observe that TS10 suffers under this protocol because all the three subjects bear similar poses for many frames. Thus, we believe the two settings are complimentary.
\indent Another evaluation metric used in MuPots Test Set is the Area Under Curve (AUC) of PCK values. We report an AUC of $35.5$ which is better than $30.1$ reported by \cite{singleshotmultiperson2018} and $27.6$ in \cite{VNect_SIGGRAPH2017} using groud truth detections. Our detection rate is $93.5\%$ which is comparable to the $93\%$ detection rate of \cite{singleshotmultiperson2018} under \textit{Setting 1}.

The above mentioned results reveal a significant increment in the state-of-the-art. It is worth noting that all the results are comparable to performance of single-person pose estimation methods.

\begin{table}[h]
\caption{Performance comparison of various training/testing settings on Human3.6M Protocol 1. The first column indicates the data used as the 2D input to the 3D pose module while training. The second column, likewise, indicates which datasets were used for training the HG-RCNN based 2D input.}\label{tab:ablation}
\begin{tabular}{| l | c | c| }
     \hline
     2D-3D Training & HG-RCNN Training  & MPJPE \\
     \hline
     H36M GT & MS-COCO & 135.5 \\
     H36M GT + noise & MS-COCO & 119.7\\
     H36M pred & MS-COCO & 73.0 \\
     H36M pred & MS-COCO + H36M & 65.2 \\
     MPI-INF GT & MS-COCO & 118.3 \\
     MPI-INF GT + noise & MS-COCO & 118.16  \\
     \hline
\end{tabular}
\vspace{-1em}
\end{table}

\textbf{Human 3.6M:} The results on Human 3.6M dataset are detailed in Table~\ref{tab: h36mp1}. We achieve an MPJPE of $65.2 mm$ after fine-tuning HG-RCNN on Human3.6M and $74.3 mm$ without fine-tuning. It may be noted that Zanfir et al.~\cite{MubyNet} report their results on the official Human3.6 test set and achieve $60mm$ MPJPE. Since the test circumstances are different, the comparison may not be fair. The combined results on MuPoTS-3D and Human3.6M also corroborate the claims in~\cite{humanMotionKanazawa19} that a good performance in Human3.6M does not necessarily indicate better generalization in wild settings.  We also evaluate our method under various test-train settings in Table.~\ref{tab:ablation} and observe that MPI-INF-3DHP~\cite{mono-3dhp2017} offers a wider range of poses to train from, thus leading to better results with ground-truth detections.
\begin{table}[h]
\centering
\caption{Comparison of HG-RCNN and Mask-RCNN based models on MuPoTS 3D. The evaluation metric is 3DPCK.}\label{tab:hgrcnn_vs_maskrcnn}
\vspace{0.5em}
\begin{tabular}{| l | c | c | }
    \hline
     & Mask-RCNN & HG-RCNN \\
     \hline
     all annotated joints & 70.1 & 72.4 \\
     all occluded joints & 61.0 & 64.1 \\
     \hline
\end{tabular}
\vspace{-1em}
\end{table}

\textbf{Mask-RCNN vs. HG-RCNN:} Table~\ref{tab:mscoco-kp} details the performance of HG-RCNN on MSCOCO Keypoints dataset. Our results are comparable to Mask-RCNN's reported results. We observe a slightly reduced mAP which can be attributed to the fact that Hourglass architecture is better suited for cases when the larger structure is to be considered. MS-COCO keypoints validation dataset contains multiple cases of isolated/truncated body parts. While evaluating on MuPoTS-3D, we observe improved 3DPCK using HG-RCNN on all annotated ($70.1\%$ vs $72.4\%$) and all occluded ($61\%$ vs $64\%$) joints alike. We also achieve comparable results on the person bounding box detections over Mask-RCNN as shown in Table~\ref{tab:mscocobb}.

\begin{table}[h]
\centering
\caption{HG-RCNN results on MS-COCO 2017 \textit{val-set} for keypoints using a ResNeXt-101 backbone.}\label{tab:mscoco-kp}
\setlength\tabcolsep{2pt}
\begin{tabular}{| l | c | c | c | c | c | }
    \hline
     & AP & AP50 & AP75 & AP\_M & AP\_L \\
     \hline
     HG-RCNN & 0.6348 & 0.8620 & 0.6905 & 0.5840 & 0.7204 \\
     \hline
\end{tabular}
\vspace{-1em}
\end{table}

\begin{table}[h]
\caption{Results on MS-COCO 2017 \textit{val-set} for person boxes.}\label{tab:mscocobb}
\setlength\tabcolsep{2pt}
\begin{tabular}{| l | c | c | c | c | c | c |}
     \hline
     & AP & AP50 & AP75 & AP\_S & AP\_M & AP\_L \\
     \hline
     HG-RCNN & 0.5536 & 0.8381 & 0.6076 & 0.3743 & 0.6320 & 0.7235 \\
     \hline
\end{tabular}
\vspace{-1em}
\end{table}

\begin{figure}[h] 
	\centering
	\includegraphics[width=1\linewidth]{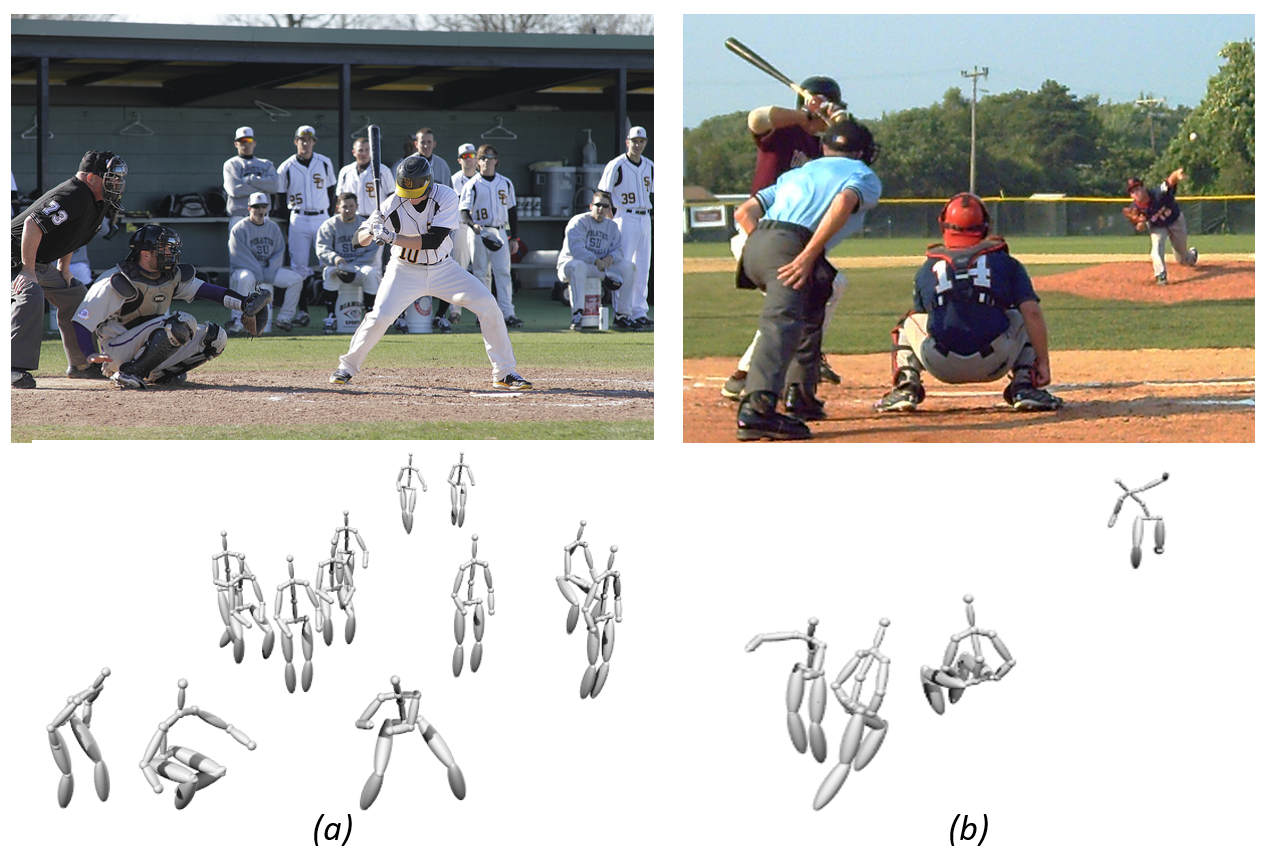}
    \caption{Two images summing up the sources of failure in our approach.} 
	\vspace{-1em}
	\label{fig:failure}
\end{figure}

\section{Limitations}
While our method attempts to account for structural information during inter-personal occlusions, we believe it can be explicitly taken care of with better structural constraints and bounding box consistencies.

\textbf{Sources of Error:} Figure~\ref{fig:failure} shows interesting examples of failure cases and exposes three sources of error in our pipeline. The first source is poor 2D keypoint estimation, which is apparent in the occluding persons of Figure~\ref{fig:failure} (b). The second source of error is an unseen activity/pose which leads to erroneous prediction. This can be seen in squatting players of both the figures, wherein the data-induced model bias leads to incorrectly predicting a person sitting on a chair instead.  

Finally, our camera-coordinate 3D pose prediction is sensitive to 2D keypoint detections and can wrongly reason about the person depth. This effect is observable in Figure~\ref{fig:failure}(a) in which the sitting people have been pushed back, in addition to the two outliers standing behind the player. It may also be noted that this approximation also assumes the individuals to be of roughly the same size. We observe incorrect relative positioning when the height difference is high. Finally, while we compute the sums of bone lengths only on the torso joints to avoid the adverse effects of fore-shortening, the effects can not be completely alleviated.

\section{Conclusion}
This paper presents a simple extension of Faster-RCNN framework to yield a near-real-time multi-person 3D human pose estimation network HG-RCNN that can be trained without a multi-person 3D pose dataset. Our proposed framework is extremely simple to implement and outperforms previous state-of-the-art results by convincing margins. We also show that we can approximate the spatial layout of the scene. These claims are substantiated both quantitatively through experimental evaluation as well as through qualitative assessments on COCO and MuPoTS-3D datasets. The paper also proposes an improvement to the greedy-matching strategy for multi-person 3D pose estimation evaluation and show results on it. In the future, we plan to deploy this pipeline to a broader human-parsing pipeline while also seeking real-life applications such as activity detection and construct a better scene understanding system related to humans. 

\section*{Acknowledgement}
This work is supported by Mercedes-Benz Research \& Development India (RD/0117-MBRDI00-001).

{\small
\bibliographystyle{ieee}
\bibliography{egbib}
}

\end{document}